\documentclass[11pt]{article}

\usepackage[final]{acl}

\usepackage{times}
\usepackage{latexsym}

\usepackage[T1]{fontenc}
\usepackage[utf8]{inputenc}

\usepackage{microtype}

\usepackage{inconsolata}

\usepackage{graphicx}

\usepackage{booktabs} % 用于实现三线表（顶部和底部的粗线 \toprule, \bottomrule）
\usepackage{multirow} % 用于实现最后一行 "SLM-SQL (Our)" 的跨行居中
\usepackage{makecell} % 用于实现同一个单元格内的手动换行（如表头和带引用的方法名）
\usepackage{xcolor}   % 用于将引用文字变成蓝色
\usepackage[table]{xcolor}
\usepackage{amssymb}
\usepackage{amsfonts}
\usepackage{placeins}

\title{VCoT-SQL: Enhancing Text-to-SQL Generation through Dual Chain-of-Thought Validation Training}

\author{
  \textbf{Rui Zhang\textsuperscript{1}},
  \textbf{Yujia Su\textsuperscript{1}},
  \textbf{Ruixuan Li\textsuperscript{1}},
  \textbf{Yuhua Li\textsuperscript{1}},
  \textbf{Yixin Su\textsuperscript{1}}\thanks{Corresponding author.}
\\
  \textsuperscript{1}
  School of Computer Science and Technology,\\
  Huazhong University of Science and Technology, Wuhan, China
\\
    \href{mailto:rayteam@yeah.net}{\texttt{rayteam@yeah.net} (\href{https://www.ruizhang.info}{www.ruizhang.info})}, \href{mailto:suyixin@hust.edu.cn}{\texttt{suyixin@hust.edu.cn}}
}

\begin{document}
\maketitle
\begin{abstract}
Open-source LLMs provide a flexible alternative to closed-source models for Text-to-SQL but still struggle with complex queries. Fine-tuning with chain-of-thought (CoT) data improves reasoning, yet standard datasets lack CoT annotations, requiring synthetic generation. Existing synthesis methods only verify the final answer, allowing flawed reasoning steps to corrupt the training data. We propose VCoT-SQL, a framework that improves Text-to-SQL generation through intermediate-step-validated CoT supervision and a dedicated SQL validator, which enables small open-source models to outperform frontier closed-source models. VCoT-SQL trains two complementary models on curated reasoning traces: an SQL generator using clause-level CoT templates where reasoning steps are verified via execution and schema alignment checks, and an SQL validator using diagnostic CoT templates to detect syntactic, semantic, and type-level errors.  We further design a hierarchical schema filtering pipeline that integrates and refines complementary retrieval strategies to distill relevant database elements. VCoT-SQL achieves 66.82\% execution accuracy on challenging dataset BIRD dev and 86.8\% on Spider test, improving the base 7B model by 7\% on BIRD. Notably, our CoT synthesis relies on 7B models yet outperforms methods that use 72B-scale or even proprietary closed-source teachers for data generation, demonstrating that rigorous step-level validation can be more effective than scaling up teacher model size. Our dataset and code are available at \href{https://github.com/INTREBID/VCoT-SQL.git}{https://github.com/INTREBID/VCoT-SQL}.

\end{abstract}

\section{Introduction}
Text-to-SQL is the task of converting natural language questions into executable SQL queries, enabling non-technical users to access and manipulate relational databases. As a fundamental problem at the intersection of natural language processing and database systems, Text-to-SQL attracts significant attention from both the academia and industry. 

LLM-based approaches dominate Text-to-SQL benchmarks, falling into two paradigms: prompt-based and fine-tuned. Prompt-based approaches primarily leverage closed-source LLMs such as GPT-4~\cite{achiam2023gpt}, which remain general-purpose models that are hard to fine-tune for Text-to-SQL, and suffer from high API costs and data privacy risks~\cite{li2024codes,pourreza2024dts,qin2025route}. Fine-tuning approaches train open-source LLMs on task-specific Text-to-SQL datasets, and recent work incorporates Chain-of-Thought (CoT)~\cite{wei2022chain} reasoning into the fine-tuning objective to strengthen reasoning capabilities~\cite{pourreza2025reasoning,zhai2025excot,li2025omnisql}. However, existing benchmarks typically provide only question–SQL pairs, and the scarcity of high-quality CoT annotations limits the effectiveness of such methods, particularly on complex queries.

Meanwhile, pipeline-based strategies are proposed to improve Text-to-SQL accuracy by augmenting the generator with schema linking~\cite{li2024codes,pourreza2024dts,qin2025route,liu2026xiyan} or Self-Correction~\cite{pourreza2023din,liu2026xiyan,talaei2024chess}. However, existing schema linking methods either require additional model training or rely on superficial semantic matching that struggles with complex schema correspondences. Moreover, self-correction approaches remain largely ineffective to semantic errors, as they focus on fixing syntax errors via execution feedback.

To address the lack of high-quality reasoning data and further improve the accuracy of generated SQL, we introduce \textbf{VCoT-SQL}, a Text-to-SQL framework built on the principle of dual validation---validating CoT quality during training and validating SQL correctness during inference. On the training side, we design dedicated CoT templates for SQL generation and SQL validation respectively, and then apply multi-stage verification (execution consistency, schema grounding, and reasoning faithfulness) to curate reliable training data for a CoT-enhanced Generator and Validator. During inference, a cascaded schema filtering pipeline first isolates relevant database elements to provide focused context; the Generator then produces candidate SQL, which the Validator diagnoses and refines by correcting syntactic errors, semantic inconsistencies, and data-type mismatches. \textbf{This design enables smaller models to achieve strong reasoning and self-correction capabilities with less data, without relying on large teacher models.}

Our contributions are as follows: 

1) We propose VCoT-SQL, a dual-validation Text-to-SQL framework with a CoT-enhanced Generator and Validator. To facilitate effective reasoning, we design dedicated CoT templates: clause-driven templates for generation and multi-type correction templates for validation, and then curate high-quality training data from Spider/BIRD via multi-stage filtering on execution consistency, schema grounding, and reasoning faithfulness.

2) We also introduce a training-free, three-stage schema retrieval framework (Pre-SQL, Column, and Value retrievers), which progressively reduces schema noise, achieving performance superior to that of trained specialized schema filters. 

3) Extensive experiments on six benchmarks validate that our method significantly improves the performance on three different base models. Our best 7B model increases execution accuracy on the BIRD dev set from 59.78\% to 66.82\%. Furthermore, it surpasses the strong OmniSQL baseline (63.9\%) while using only \textbf{1/250} of its training data.

\section{Related Work}

\subsection{Text-to-SQL}
Early Text-to-SQL systems rely on rule-based methods~\cite{woods1972lunar,li2014nalir}, which struggle with complex and cross-domain queries. Large-scale benchmarks such as WikiSQL~\cite{zhong2017seq2sql} and Spider~\cite{yu2018spider} have spurred neural approaches~\cite{zhong2017seq2sql,yu2018syntaxsqlnet,guo2019towards}, yet these methods still struggle to generalize to unseen schemas and complex query intents.

Pretrained language models such as BERT~\cite{devlin2019bert} and T5~\cite{raffel2020exploring} bring contextual representations to Text-to-SQL~\cite{hwang2019comprehensive,qi2022rasat,li2023resdsql,li2023graphix}. With the rise of large language models (LLMs), current approaches broadly fall into two paradigms. Prompt-based methods leverage powerful closed-source LLMs via in-context learning~\cite{pourreza2023din,gao2024texttosql,pourreza2024chase,talaei2024chess,lee2025mcs}, but incur high inference costs, while also precluding task-specific fine-tuning. Conversely, fine-tuning strategies tailor LLMs to the task, aiming to bolster structural robustness and execution fidelity~\cite{pourreza2024dts,yang2024synthesizing,li2024codes}. However, constrained by their limited model capacity, these approaches often exhibit insufficient reasoning ability on complex queries. 

Schema linking in Text-to-SQL is a type of retrieval-augmented generation, as it retrieves relevant schema elements to guide SQL generation. For a comprehensive survey on multimodal retrieval-augmented generation including the modality of tables and schemas, please refer to~\citet{zhang2025MM-RAGsurvey}. However, existing schema
linking methods typically rely on a single strategy, such as model
training~\cite{pourreza2024dts,li2024codes,qin2025route}, embedding
matching~\cite{liu2026xiyan}, or LLM-based
extraction~\cite{lee2025mcs}. Meanwhile, existing self-correction approaches
rely either on prompt-based methods~\cite{pourreza2023din} or on
execution-error-driven correctors without CoT-enhanced
training~\cite{liu2026xiyan,qin2025route}.

\subsection{Chain-of-Thought Fine-tuning}

Chain-of-Thought (CoT) reasoning has proven effective for complex tasks. Recent models such as OpenAI o1~\cite{jaech2024openai} and DeepSeek-R1~\cite{guo2025deepseek} internalize it through specialized training objectives and reinforcement learning. A key challenge in CoT fine-tuning is acquiring high-quality reasoning traces. Since manual annotation is costly to scale~\cite{chung2024scaling}, subsequent work synthesizes CoT data via LLMs with filtering heuristics~\cite{zelikman2022star,kim2023cot}. For Text-to-SQL, OmniSQL~\cite{li2025omnisql} collects a large-scale CoT-enriched dataset via free-form prompting. However, such approaches~\cite{pourreza2025reasoning,zhai2025excot,li2025omnisql} depend on the capabilities of ultra-large models and rely solely on execution correctness as the filter, without validating the reliability of intermediate reasoning steps.

\begin{figure*}
  \centering
  \includegraphics[width=\linewidth]{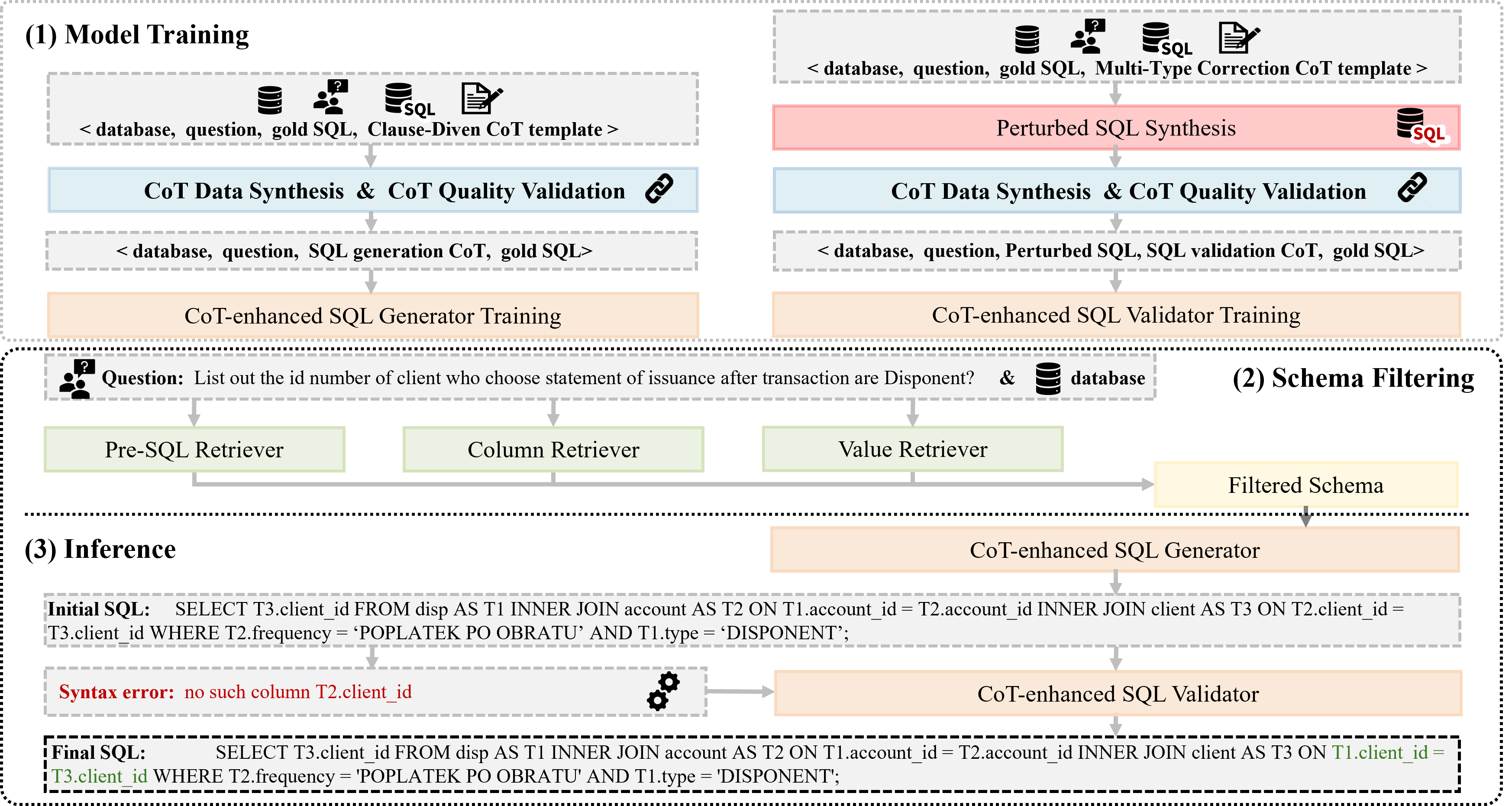}
  \caption{Overview of the proposed VCoT-SQL framework. (1) \textbf{Model Training}: two pipelines collect CoT supervision data to train the CoT-enhanced SQL Generator and Validator. (2) \textbf{Schema Filtering}: the question and full database are processed through Pre-SQL, Column, and Value Retrieval to obtain a filtered schema (see Figure~\ref{schema_serialization}). (3) \textbf{Inference}: the Generator produces an initial SQL from the filtered schema; the SQL is executed, and any syntax errors along with the initial SQL are passed to the Validator to yield the final SQL.}
   \label{fig1}
\end{figure*}

\section{Methodology}

As illustrated in Figure~\ref{fig1}, our VCoT-SQL can be divided into three stages. In the model training stage, we synthesize and validate CoT data based on the designed clause-driven CoT template to train the CoT-enhanced SQL Generator (\S\ref{sec:method_intro1}); similarly, we train a CoT-enhanced SQL Validator based on the multi-perspective correction CoT (\S\ref{sec:method_intro2}).  In the schema filtering stage, complementary Pre-SQL Retriever, Column Retriever, and Value Retriever are applied to progressively simplify the schema. In the inference stage, the SQL Generator produces a candidate SQL based on the filtered schema, which is then fed into the SQL Validator for multi-faceted correction to yield the final SQL (\S\ref{sec:method_intro4}).

\subsection{CoT-enhanced SQL Generator Training}
\label{sec:method_intro1}

\textbf{Clause-Driven CoT}
To guide structured reasoning for SQL generation, we design a clause-driven CoT template that mirrors the inherent structure of SQL, decomposing the task into clause-level reasoning steps, including the SELECT, WHERE, FROM, and other clauses. Each clause is further divided into fine-grained steps that specify how to identify relevant columns, infer filtering conditions, determine join relationships, and construct aggregations. Interestingly, although SQL queries typically follow the SELECT–FROM–WHERE order, we find that reasoning about the FROM clause only after analyzing the others produces higher-quality CoT and more accurate SQL. This reordering allows the model to select tables only after identifying the required columns, conditions, and aggregations, thereby reducing spurious joins and unnecessary tables. Overall, this structured CoT format aligns with the hierarchical logic of SQL while enabling reliable programmatic verification of reasoning quality. 

\textbf{CoT Data Synthesis \& CoT Quality Validation}
Based on the clause-structured template, we construct CoT data using few-shot strategy. We provide three high-quality CoT exemplars, manually crafted from samples in Spider and BIRD, as references for the LLM. For each training instance, we also provide its gold SQL to improve generation accuracy. However, since a small portion of the original dataset contains mismatches~\cite{pourreza2025reasoning} between questions and gold SQL, we instruct the LLM to first verify whether the provided SQL is correct and then generate structured CoT reasoning steps followed by the final SQL output. The complete CoT template and the prompt for CoT generation are presented in Appendix~\ref{appendix:sql_generation_prompt}.

Although providing the gold SQL upfront during CoT synthesis can alleviate part of the generation burden, we observe that this sometimes leads to cases where the final SQL is correct yet the intermediate reasoning steps are unreliable or unsupported. To address this, we introduce a multi-stage verification mechanism evaluating both execution consistency and schema grounding. We first filter out CoT data that violates the expected clause structure or exceeds a length threshold. Next, we check whether the SQL derived from the CoT executes to the same result as the gold SQL. Finally, and most critically, we validate the CoT reasoning itself. We observe that providing gold SQL during generation can cause answer leakage, where the reasoning references phrases such as 'Gold SQL'; such instances are removed. We then extract schema elements from the gold SQL and verify via heuristic rule-based matching that they are appropriately referenced in the CoT, ensuring the reasoning correctly grounds on the relevant schema.

\textbf{Generator Training}
We fine-tune the SQL generator via supervised learning. The loss function is defined as follows: 
\begin{equation}
\mathcal{L}_{gen} = -\mathbb{E}_{(x,y)\sim D_{gen}} \left[ \sum_t \log P_{\theta_{gen}}(y_t \mid x, y_{<t}) \right]
\end{equation}
where $D_{gen}$ represents the training set for the generator. The input sequence $x$ consists of the user question and the database schema. The output sequence $y$ comprises the validated clause-driven reasoning followed by the generated SQL. The generator is optimized to predict the CoT solution and the corresponding SQL based on the provided inputs.
  
\subsection{CoT-enhanced SQL Validator Training}
\label{sec:method_intro2}
\textbf{Multi-Type Correction CoT}
We adopt a structured CoT-driven correction framework that guides the LLM to diagnose and repair common SQL error types \cite{shen2025study,xu2025boosting}. For syntactic correction, execution-time errors from the database engine are provided as diagnostic evidence. For semantic correction, the model evaluates table–column usage, filtering conditions, and correctness of GROUP BY/ORDER BY/LIMIT clauses. We further incorporate data-type verification, guiding the model to check for issues such as proper numeric casting to preserve precision and correct date formatting. Through this multi-type reasoning process, the LLM incrementally refines its prediction into a query that is both executable and semantically faithful to the question intent.

\textbf{Perturbed SQL Synthesis}
To construct training data for SQL validation, we need imperfect SQL queries, as original training sets contain only gold SQL. We generate errors by prompting an LLM to systematically perturb the gold SQL, using exclusively those passing CoT validation during the SQL generation stage to largely ensure semantic alignment with the question. To align synthesized errors with the types handled by our validation template, we organize common SQL mistakes into several categories: column-related, table-related, condition-related, aggregation-related and data-type misuses such as improper numerical casting or incorrect date handling. For each category, we design a concise instruction that prompts the LLM to rewrite the SQL with the targeted defect.

\textbf{CoT Data Synthesis \& CoT Quality Validation}
After obtaining erroneous–gold SQL pairs, we synthesize validation-oriented CoT following the same procedure as in the SQL generation setting. A small set of handcrafted, high-quality CoT exemplars from real SQL errors serves as few-shot demonstrations. More detailed exemplar design and the complete CoT template for SQL validation are presented in Appendix~\ref{appendix:v_exampler} and Appendix~\ref{appendix:sql_validation_prompt}.

For the correction CoT, we first filter out samples that violate the expected format or exceed a length threshold, and remove those explicitly mentioning "Gold SQL". We then check execution consistency between the final corrected SQL in the CoT and the gold SQL, discarding mismatched instances. Finally, given the complex yet well-structured nature of the correction reasoning, we validate its faithfulness by extracting only the analysis and correction steps from the CoT, feeding them to the LLM, and verifying that the regenerated corrected SQL is consistent with the gold SQL. This ensures the reasoning process itself is reliable, not merely producing correct results by coincidence.

\textbf{Validator Training}
Following the similar paradigm as Generator, we leverage high-quality SQL-validation CoT data to train the validator. The loss function is defined as follows: 
\begin{equation}
\mathcal{L}_{val} = -\mathbb{E}_{(x,y)\sim D_{val}} \left[ \sum_t \log P_{\theta_{val}}(y_t \mid x, y_{<t}) \right]
\end{equation}
where $D_{val}$ represents the training set. The input sequence $x$ consists of the question, database schema, generated SQL and syntax error (null is correct). The output sequence $y$ comprises the validated correction reasoning followed by the validated SQL.

\subsection{Schema Filtering and Inference}
\label{sec:method_intro4}
Large-scale databases often contain redundant schema elements that introduce noise. Prior methods such as CodeS~\cite{li2024codes} address this by training dedicated schema filter models. We propose a training-free schema retrieval framework (Pre-SQL Retriever, Column Retriever, and Value Retriever) that jointly leverages LLM's code-generation capability and lightweight semantic encoding to progressively prune schema.

\textbf{Pre-SQL Retriever}
The first retrieval stage leverages an LLM to generate a coarse-grained Pre-SQL based on the question and complete schema containing only table names, column names, and types. We parse the Pre-SQL to extract tables, columns and literal values, which also provide valuable cues for schema filtering.

\textbf{Column Retriever}
To mitigate omissions from the Pre-SQL, we further apply an embedding-based column retrieval step. We encode both the full question and LLM-extracted fine-grained keywords into semantic embeddings, and match them against all database column representations to retrieve the top-K candidates. The full-question embedding captures global context, while keyword embeddings provide complementary fine-grained signals.

\textbf{Value Retriever}
Some user queries implicitly reference database values without mentioning corresponding columns. Inspired by CHESS~\cite{talaei2024chess}, the Value Retriever employs a three-stage filtering pipeline—LSH indexing, semantic similarity, and edit distance—to identify hidden dependencies. We query the value index using literals from the Pre-SQL and LLM-extracted keywords.

\textbf{Inference}
Finally, we take the union of all columns retrieved from the three paths and serialize the filtered schema into a compact representation enriched with full table names, nullable indicators, and matched values. During inference, the Generator first produces an initial SQL query, which is executed to collect any syntax errors (empty if correct). The Validator then takes the query and error feedback to perform syntactic, semantic, and data-type diagnostics, producing the refined output. Further details are in Appendix~\ref{appendix:schema}.

\begin{table*}[t]
\centering
\small

\setlength{\tabcolsep}{4pt}
\renewcommand{\arraystretch}{1.0}
\begin{tabular}{l|c|c|cccccc}
\toprule
\makecell{Method} & 
\makecell{Model} & 
\makecell{Training  \\ Data Size} & 
\makecell{Spider \\ Dev} & 
\makecell{Spider \\ Test} & 
\makecell{BIRD \\ Dev} & 
\makecell{Spider- \\ DK} & 
\makecell{Spider- \\ Syn} & 
\makecell{Spider- \\ Realistic} \\
\midrule

% --- 区块 1：闭源大模型基线 ---
\multicolumn{8}{c}{Prompting with Closed-Source LLMs} \\

\midrule
\makecell[l]{GPT-4o \\ {~\cite{openai_hello_2024}}}      &
- & 
- & 
81.9 & 
83.0 & 
61.93 & 
70.8 & 
74.2 & 
82.9 \\

\makecell[l]{GPT-4o-mini \\ {~\cite{openai_gpt-4o_2024}}} & 
- & 
- & 
82.9 & 
82.8 & 
60.43 & 
73.6 & 
74.2 & 
81.7\\

\makecell[l]{GPT-4-turbo \\ {~\cite{openai_hello_2024}}} & 
- & 
- & 
83.8 & 
85.2 & 
62.06 & 
71.8 & 
77.5 & 
83.2\\
\midrule

\multicolumn{8}{c}{Prompting with Open-Source LLMs} \\
\midrule
% --- 区块 2：开源大模型基线 ---
\makecell[l]{Llama3 \\{~\cite{touvron2023llama}}}   
& Llama3-8B
& -
& 69.3 
& 69.1 
& 39.05 
& 57.2 
& 62.6 
& 72.8 \\

% 修改点 1：使用 \rowcolor 给整行上色，并在第一列使用 \cellcolor{white} 覆盖为白色
\makecell[l]{DeepSeek-Coder \\ {~\cite{guo2024deepseek}}}  
& DeepSeek-Coder-6.7B-Instruct 
& -
& 72.4 
& 72.3 
& 49.54 
& 62.4 
& 61.8 
& 68.3\\

\makecell[l]{Qwen2.5 \\{~\cite{team2024qwen2}}}  
& Qwen2.5-7B-Instruct 
& -   
& 77.9
& 78.2 
& 53.98 
& 68.0 
& 67.7 
& 76.0 \\

\makecell[l]{Qwen2.5-Coder \\{~\cite{hui2024qwen2}}} 
& Qwen2.5-Coder-7B-Instruct 
& -  
& 82.3 
& 81.0 
& 58.87 
& 70.9 
& 71.3 
& 80.9\\

\makecell[l]{XiYanSQL \\ {~\cite{liu2026xiyan}}}       
& XiYanSQL-QwenCoder-7B 
& - 
& 86.9 
& 84.7 
& 59.78 
& 73.1 
& 75.7 
& 77.6 \\
\midrule

\multicolumn{8}{c}{Fine-Tuning with Open-Source LLMs} \\
\midrule
% --- 区块 3：现有 Text-to-SQL 框架 ---

\shortstack[l]{CodeS \\ {~\cite{li2024codes}}}        
& StarCoder 7B
& \shortstack{ 18K } 
& 85.4 
& 83.3 
& 57.17 
& 66.9 
& 73.8 
& 81.5 \\

\shortstack[l]{ROUTE \\ {~\cite{qin2025route}}}    
& Llama3-8B   
& \shortstack{ 18K } 
& 86.0 
& 83.9 
& 57.30 
& 74.6 
& 77.4
& 80.9\\

\shortstack[l]{OmniSQL \\ {~\cite{li2025omnisql}}}     
& Qwen2.5-Coder-7B-Instruct
& \shortstack{2.5M} 
& 86.2 
& \textbf{87.9}
& 63.90 
& 76.1 
& 77.3 
& 81.9\\
\midrule

% --- 区块 4：我们的方法 ---
% 修改点 2：同样使用 \rowcolor 和 \cellcolor{white}，注意 \cellcolor{white} 放在 \multirow 前面
\textbf{\shortstack{Ours: SQL Generator}} 
&  \makecell[c]{XiYanSQL-QwenCoder-7B}   
& \shortstack{9K}
& \underline{87.2} 
& 86.4
& \underline{65.38}
& \underline{76.6} 
& \underline{80.0} 
& \underline{84.3}\\

\midrule

\textbf{Ours: VCoT-SQL}
&  \makecell[c]{XiYanSQL-QwenCoder-7B}   
& \shortstack{10K}
& \textbf{88.0} 
& \underline{86.8}
& \textbf{66.82}
& \textbf{76.8} 
& \textbf{80.4} 
& \textbf{85.4}\\
\bottomrule

\end{tabular}
\caption{Performance of Text-to-SQL Models and Methods. To ensure fairness in comparison, we used the results from the greedy decoding released officially by OmniSQL. SQL Generator denotes single-pass generation; VCoT-SQL further applies our SQL Validator. For Training Data Size, prompting-based methods require no Text-to-SQL data (marked as "-"). Among fine-tuning methods, 18K refers to the combined Spider and BIRD training sets, 2.5M to OmniSQL's SynSQL-2.5M, and our 9K/10K to CoT data constructed from the Spider and BIRD training splits.}
\label{tab:main_results}
\end{table*}

\section{Experiments}

\subsection{Experiments Setting}

\textbf{Evaluation Dataset}
We evaluate VCoT-SQL on two widely used benchmarks, Spider~\cite{yu2018spider} and BIRD~\cite{li2023can}, along with three additional datasets, Spider-Syn~\cite{gan2021towards}, Spider-Realistic~\cite{deng2021structure} and Spider-DK~\cite{gan2021exploring}. We use Execution Accuracy (EX) \cite{yu2018spider} as our evaluation metric with the official scripts to ensure fair comparisons with prior work.

\textbf{Baselines}
Given that VCoT-SQL is a fine-tuned open-source model, we primarily compare it with other open-source models (Qwen2.5~\cite{team2024qwen2}, Qwen2.5-Coder~\cite{hui2024qwen2}, DeepSeek-Coder~\cite{guo2024deepseek}, XiYan-SQL~\cite{liu2026xiyan}, Llama3~\cite{touvron2023llama}), particularly those fine-tuned for Text-to-SQL like CodeS~\cite{li2024codes}, ROUTE~\cite{qin2025route}, OmniSQL~\cite{li2025omnisql}. Additionally, to position our work within the broader landscape, we compare against top-performing closed-source models (GPT-4o-mini~\cite{openai_gpt-4o_2024}, GPT-4-Turbo~\cite{openai_gpt-4_2024}, GPT-4o~\cite{openai_hello_2024}). Unless otherwise specified, our Schema Filter is adopted as the default input preprocessing step for all prompting-based and our methods to reduce input length and ensure consistent evaluation. More details can be found in Appendix \ref{appendix:im_details}.

\subsection{Main Results}

\begin{table*}
\centering
\small
\setlength{\tabcolsep}{5.5pt}
\renewcommand{\arraystretch}{1.0}
\begin{tabular}{l|cccc|cc|cccccc}
\toprule
\multirow{2}{*}{Base Model} & 
\multicolumn{2}{c}{\makecell{Spider \\ Dev}} & 
\multicolumn{2}{c|}{\makecell{Spider \\ Test}} & 
\multicolumn{2}{c|}{\makecell{BIRD \\ Dev}} & 
\multicolumn{2}{c}{\makecell{Spider- \\ DK}} & 
\multicolumn{2}{c}{\makecell{Spider- \\ Syn}} & 
\multicolumn{2}{c}{\makecell{Spider- \\ Realistic}} \\
& EX & $\Delta$ & EX & $\Delta$ & EX & $\Delta$ & EX & $\Delta$ & EX & $\Delta$ & EX & $\Delta$ \\
\midrule

% --- 区块 1：仅生成 ---
\multicolumn{12}{l}{\textbf{Only SQL Generator}} \\
\midrule
\makecell[c]{DeepSeek-Coder-6.7B-Instruct} & 84.9 & - & 86.9 & - & 60.63 & - & 74.6& -  & 77.5& -  & 80.1& -  \\
\makecell[c]{Qwen2.5-Coder-7B-Instruct}    & 85.3 & - & 86.7& -  & 64.86& -  & 76.8 & - & 77.6 & - & 82.7& -  \\
\makecell[c]{XiYanSQL-QwenCoder-7B}   & 87.2 & - & 86.4 & -  & 65.38 & - & 76.6 & - & 80.0 & - & 84.3 & - \\
\midrule

% --- 区块 2：基础模型验证 ---
\multicolumn{12}{l}{\textbf{w/  Base Model Validation}} \\
\midrule
\makecell[c]{DeepSeek-Coder-6.7B-Instruct} & 80.1 & -4.8  & 80.3 & -6.6 & 56.06 & -4.57 & 69.2& -5.4 & 69.5& -8.0 & 75.6& -4.5 \\
\makecell[c]{Qwen2.5-Coder-7B-Instruct}    & 85.5& +0.2 & 86.9 & +0.2 & 65.91 & +1.05 & 75.9 & -0.9 & 77.8 & +0.2 & 82.9 & +0.8 \\
\makecell[c]{XiYanSQL-QwenCoder-7B}   & 87.7 & +0.5 & 86.7 & +0.3 & 65.71 & +0.33  & 76.6 & - & 80.5 & +0.5  & 84.6 & +0.3\\
\midrule

% --- 区块 3：SFT模型验证 ---
\multicolumn{12}{l}{\textbf{w/  SQL Validator}} \\
\midrule
\makecell[c]{DeepSeek-Coder-6.7B-Instruct} & 85.1& +0.2 & 87.1& +0.2 & 61.41& +0.78 & 75.1 & +0.5 & 77.6 & +0.1 & 80.3 & +0.2 \\
\makecell[c]{Qwen2.5-Coder-7B-Instruct}    & 85.7 & +0.4 & 86.9& +0.2 & 66.49& +1.63 & 77.2 & +0.4 & 78.5 & +0.9 & 83.3 & +0.6 \\
\makecell[c]{XiYanSQL-QwenCoder-7B}   & 88.0& +0.8 & 86.8 & +0.4 & 66.82 & +1.44 & 76.8 & +0.2 & 80.4 & +0.4 & 85.4 & +1.1\\
\bottomrule
\end{tabular}
\caption{Ablation study on SQL validation strategies across three backbones. EX denotes execution accuracy (\%) and $\Delta$ indicates absolute improvement over the generation-only baseline.}
\label{tab:SQL_Verification}
\end{table*}

\textbf{BIRD Results}
BIRD is challenging due to its complex schemas and SQL structures. As shown in Table \ref{tab:main_results}, VCoT-SQL achieves state-of-the-art performance among models of comparable size, reaching 66.82\% EX. With our dual-module architecture, VCoT-SQL generates high-quality reasoning and corrects potential errors, surpassing even GPT-4-turbo (62.06\%) and GPT-4o (61.93\%).

Notably, both VCoT-SQL and OmniSQL~\cite{li2025omnisql} excel beyond direct-tuning methods e.g. ROUTE~\cite{qin2025route}, CodeS~\cite{li2024codes}, showing the importance of explicit reasoning for complex SQL generation. Compared with OmniSQL, VCoT-SQL achieves a 2.92\% improvement using only 10K validated CoT samples, and even our single-module version (65.38\%) surpasses OmniSQL, demonstrating the effectiveness of our multi-stage CoT validation.

\textbf{Spider and Its Variants Results}
We further evaluate VCoT-SQL on Spider and its robustness variants (Spider-DK, Spider-Syn, Spider-Realistic). As shown in Table \ref{tab:main_results}, VCoT-SQL consistently improves over its XiYanSQL-QwenCoder-7B backbone across all benchmarks: 88.0\% vs. 86.9\% on Spider Dev, 86.8\% vs. 84.7\% on Test, and similar gains on robustness variants. OmniSQL achieves higher EX on Spider Test (87.9\%), likely benefiting from its larger 2.5M-sample SynSQL training corpus. However, VCoT-SQL demonstrates superior data efficiency, achieving competitive performance with only 10K validated samples. 

\textcolor{black}{We further evaluate robustness on all 17 perturbation test sets of Dr.Spider~\cite{chang2023drspider}. The complete results and category-level analysis are reported in Appendix~\ref{app:drspider}.}

\subsection{Ablations on SQL Validation}

\begin{table*}[t]
\centering
\small
\setlength{\tabcolsep}{5pt}
\renewcommand{\arraystretch}{1.1}
\begin{tabular}{l|cc|c|ccc}
\toprule
\makecell{Method} & 
\makecell{Spider \\ Dev} & 
\makecell{Spider \\ Test} & 
\makecell{BIRD \\ Dev} & 
\makecell{Spider- \\ DK} & 
\makecell{Spider- \\ Syn} & 
\makecell{Spider- \\ Realistic} \\
\midrule

% --- 第1行 ---
Base Model (Qwen2.5-Coder-7B-Instruct)
&  82.3 & 81.0 & 58.87 & 70.9 & 71.3 & 80.9 \\
\midrule

% --- 第2行 ---
\makecell[l]{FT w/ original Spider and BIRD} 
&  82.6 & 83.1 & 60.04 & 73.6 & 74.5 & 79.5 \\

\makecell[l]{FT w/ unvalidated CoT-enhanced (Syntax-Order) Spider and BIRD}
&
83.8
&
84.1
&
60.69
&
73.3
&
74.9
&
80.0\\

% --- 第3行 ---
\makecell[l]{FT w/ unvalidated CoT-enhanced (Logic-Order) Spider and BIRD} 
&  84.4 & 84.9 & 61.60 & 73.5 & 75.0 & 79.9 \\

% --- 第4行 ---
\makecell[l]{FT w/ validated CoT-enhanced (Syntax-Order) Spider and BIRD} 
& 85.0 & 86.8 & 62.58 & 76.3 & 77.2 & 82.3 \\

% --- 第5行 ---
\makecell[l]{FT w/ validated CoT-enhanced (Logic-Order) Spider and BIRD} 
&  85.3 & 86.7 & 64.86 & 76.8 & 77.6 & 82.7 \\
\bottomrule

\end{tabular}
\caption{Ablation Results for Validated SQL Generation on different data construction settings (EX).}
\label{tab:CoT_Synthesis}
\end{table*}

To assess our SQL Validator and the necessity of task-specific fine-tuning, we conduct an ablation study across three base models. As shown in Table \ref{tab:SQL_Verification}, we compare standalone generation against pipelines using either a base or our fine-tuned model for validation. Interestingly, simply appending a base model to verify generated SQL yields highly unstable and often detrimental results. For instance, using base DeepSeek-Coder leads to a 4.57\% absolute decline on BIRD Dev, with similar drops observed on other datasets. Further analysis reveals clear model-dependent behavior: compared with other backbones, DeepSeek-Coder tends to regenerate SQL from scratch rather than preserving and correcting existing queries, introducing more false-positive edits and degrading performance. This indicates that without task-specific alignment, raw LLMs lack the constrained SQL debugging capabilities.

In contrast, our fine-tuned validator consistently improves execution accuracy across all backbones and benchmarks. Trained on validated CoT data, the model acquires effective error detection and correction capabilities. On BIRD Dev, the SFT validator boosts Qwen2.5-Coder by 1.63\%, with similar gains on other baselines. Among initially incorrect SQLs, the Validator successfully corrects 1.83\% of cases, while among already correct queries, only 0.20\% are incorrectly altered. These results demonstrate our validator's reliability and effectiveness. Further analyses demonstrating the stability of the validator are presented in Appendix~\ref{appendix:validator}.

\textcolor{black}{We also compare our fine-tuned 7B Validator with GPT-5.4 under the same generator outputs and evaluation protocol. We treat this frontier model only as an empirical upper-bound reference; results are provided in Appendix~\ref{app:frontier_validator}.}

\subsection{Ablations on Validated SQL Generation}

To evaluate the effectiveness of our CoT quality validation strategy for SQL generation, we conduct ablation experiments using the same backbone under five training data settings: original Spider and BIRD; unvalidated and validated CoT data generated with the syntax-order template (\S\ref{sec:method_intro1}); unvalidated and validated CoT data generated with our logic-order template (used in the final model).

As shown in Table \ref{tab:CoT_Synthesis}, training with synthesized CoT data already improves performance over the original training data, increasing accuracy from 58.87\% to 61.60\% on BIRD Dev and from 81.0\% to 84.9\% on Spider Test, demonstrating the general benefit of CoT for SQL generation. Applying our CoT quality validation boosts performance by filtering flawed reasoning. Among all settings, validated logic-order CoT achieves the best results, reaching 64.86\% on BIRD Dev. This represents a consistent improvement of 2–6\% over the base model fine-tuned directly on original datasets, highlighting the broad effectiveness of our structured reasoning supervision. Compared with syntax-order, the logic-order performs more consistently, suggesting that organizing reasoning according to query planning is more effective than SQL syntax order. \textcolor{black}{To further disentangle the effects of template ordering and CoT validation, we conduct a 2$\times$2 factorial analysis over the four corresponding settings. Detailed results are provided in Appendix~\ref{app:factorial}.}

\subsection{Ablations on Schema Filter}
\begin{table*}[t]
\centering
\small
\renewcommand{\arraystretch}{1.1}
\setlength{\tabcolsep}{4.5pt}
\begin{tabular}{l|ccc|ccc|ccc}
\toprule
\multirow{2}{*}{\makecell{Method}} & 
\multicolumn{3}{c|}{\makecell{Spider Dev}} & 
\multicolumn{3}{c|}{\makecell{Spider Test}} & 
\multicolumn{3}{c}{\makecell{BIRD Dev}} \\
& Precision & Recall  & EX & Precision & Recall  & EX & Precision & Recall & EX \\
\midrule

Full schema        & - & -  & 83.6 & - & - & 86.6 & - & - & 59.71  \\
\midrule

CodeS schema filter    & 18.02 & 99.83   & 85.4 & 20.22 & 99.87  & 85.1 & 16.44 & 97.76 & 62.63  \\
\midrule

\textbf{VCoT-SQL schema filter}  & 30.85 & 99.79  & 85.3 & 34.51 & 99.89 & 86.7 & 31.36& 98.09 & 64.86  \\
w/o Pre-SQL retriever & 30.77 & 98.94 & 83.8& 33.92&97.74 & 84.4 & 31.39 & 94.91 &61.15  \\
w/o column retriever & 67.35& 97.02& 83.6&70.02&96.26& 84.5 &65.48& 92.30& 61.08 \\
w/o value retriever  &30.47 & 99.77  & 85.0 & 34.14 & 99.82& 86.6 &31.37&97.84 & 63.23 \\
\bottomrule

\end{tabular}
\caption{Comparison and ablation results of the three-path schema retrieval framework. Precision and Recall assess the discrepancy between the retrieved schema and the gold schema (only the tables and columns appearing in the gold SQL). EX evaluates the downstream generation accuracy of our Generator (Qwen2.5-Coder-7B-Instruct).}
\label{tab:Schema_Filter}
\end{table*}

\begin{table}[t]
\centering
\small
\renewcommand{\arraystretch}{0.8}
\begin{tabular}{l|c|c|c}
\toprule
\makecell{Qwen2.5-Coder\\-7B-Instruct}& \makecell{Full \\ Schema} & \makecell{Schema\\ Filter} & \makecell{$\Delta$ Schema \\Filter} \\
\midrule

% --- EXCoT ---
Base Model & 53.92&	58.87&	+4.95 \\
\midrule

% --- OMNI-SQL ---
Dual-Validation & 61.34	&66.49	&+5.15\\
\midrule

% --- CoT^2-SQL ---
$\Delta$ Dual-Validation & +7.42&	+7.62	&— \\
\bottomrule

\end{tabular}
\caption{2×2 ablation of Dual-Validation and Schema Filter on BIRD Dev (EX).}
\label{tab:ablition} 
\end{table}

We evaluate our three-path schema filter against the full schema and the trained schema filter from CodeS. As shown in Table \ref{tab:Schema_Filter}, our method improves retrieval quality across all datasets, increasing Precision while maintaining near-complete Recall. For example, Precision improves from 16.44\% (CodeS) to 31.36\% on BIRD Dev, while Recall remains above 98\%. This indicates that our method effectively removes irrelevant schema elements without sacrificing critical schema coverage, which further translates into stronger and more stable downstream EX performance. To investigate the contribution of each retrieval component, we further ablate each retrieval path individually. Removing either the Pre-SQL retriever or the Value retriever degrades both retrieval quality and EX, confirming their roles in structural grounding and value-aware schema matching. Removing the Column Retriever increases Precision but reduces Recall and EX, indicating that although broader retrieval introduces redundancy, it helps ensure coverage of necessary schema elements for SQL generation. This suggests that missing critical schema elements is more harmful than introducing moderate redundancy.

We further conduct a 2×2 ablation on the more challenging BIRD Dev dataset using Qwen2.5-Coder-7B-Instruct as the backbone to analyze the contributions of Dual-Validation (including CoT data quality validation and SQL generation validation) and Schema Filter. As shown in Table \ref{tab:ablition}, Schema Filter consistently improves performance under both settings (+4.95 / +5.15), while Dual-Validation yields larger gains (+7.42 / +7.62). This indicates that Dual-Validation is the primary source of improvement, while Schema Filter also provides a non-negligible complementary contribution.

\begin{table}[t]
\centering
\small
\renewcommand{\arraystretch}{0.8}
\begin{tabular}{l|l|cc|c}
\toprule
\multirow{2}{*}{Method} & 
\multirow{2}{*}{\makecell{Data Synthesis \\ Model}} &  
\multicolumn{2}{c|}{Spider} & 
\multicolumn{1}{c}{BIRD} \\
& & Dev & Test & Dev \\
\midrule

% --- EXCoT ---
EXCoT & 
\makecell[c]{GPT-4o} &
- & 81.2 & 59.65 \\
\midrule

% --- OMNI-SQL ---
OmniSQL & 
\makecell[c]{Qwen2.5-Coder-\\32B-Instruct \\ Qwen2.5-\\72B-Instruct} &
80.3 & 86.6 & 59.60 \\
\midrule

% --- CoT^2-SQL ---
VCoT-SQL & 
\makecell[c]{Qwen2.5-Coder-\\7B-Instruct \\ Qwen2.5-\\7B-Instruct} & 
\textbf{85.3} & \textbf{86.7} & \textbf{64.86} \\
\midrule
\textcolor{black}{VCoT-SQL} & 
\textcolor{black}{\makecell[c]{Qwen2.5-Coder-\\32B-Instruct \\ Qwen2.5-\\72B-Instruct}} & 
\textcolor{black}{\textbf{86.2}} & \textcolor{black}{\textbf{87.3}} & \textcolor{black}{\textbf{65.06}} \\
\bottomrule

\end{tabular}
\caption{Data synthesis setting and downstream performance (EX). ExCoT and OmniSQL report results trained on their synthesized CoT data from the Spider and BIRD training sets, VCoT-SQL follows the same setting. For the last row, the downstream Generator is fixed to Qwen2.5-Coder-7B-Instruct; only the data synthesis models are scaled.}
\label{tab:Model_Evolution} 
\end{table}
\section{Analysis at Efficiency of Data Synthesis}

Beyond final execution accuracy, a key advantage of VCoT-SQL is the efficiency of its data synthesis pipeline. Existing methods rely on heavy external resources. For example, OmniSQL uses a brute-force scaling strategy requiring ensembles of large models (32B and 72B) for knowledge distillation. Similarly, EXCoT~\cite{zhai2025excot} adopts a divide-and-conquer CoT template~\cite{pourreza2024chase} and relies on GPT-4o for synthetic SFT data. In contrast, the entire VCoT-SQL pipeline is self-driven by 7B-parameter models.

Under the premise that both methods use only the Spider and BIRD training sets as foundational data, Table ~\ref{tab:Model_Evolution} shows that despite relying on smaller teacher models, VCoT-SQL achieves superior downstream performance. Compared to OmniSQL, our method delivers an absolute improvement of 5.26\% on BIRD Dev and 5.0\% on Spider Dev, while maintaining competitive performance on Spider Test. It also surpasses EXCoT on BIRD Dev despite not relying on GPT-4o-generated data. \textcolor{black}{Scaling the data synthesis models to 32B/72B improves VCoT-SQL to 86.2\% on Spider Dev, 87.3\% on Spider Test, and 65.06\% on BIRD Dev, while the downstream Generator remains fixed at 7B.} These results suggest that scaling model size is not the only path to stronger Text-to-SQL reasoning. With a structured CoT template and strict validation, high-quality reasoning trajectories can be bootstrapped by 7B-class models. This "quality over quantity" paradigm reduces dependence on expensive large models and massive datasets.

\section{Conclusion}
This paper introduces VCoT-SQL, a generalizable Text-to-SQL framework that enhances SQL generation through dual validation by training on validated chain-of-thought data and validating generated SQL. We design tailored CoT templates and CoT quality validation mechanisms, enabling the collection of high-quality training data even with smaller models. Through supervised training on two specialized datasets constructed with this approach, we obtain the CoT-enhanced generator and validator. Extensive experiments on Spider, BIRD, and related benchmarks demonstrate that VCoT-SQL achieves competitive execution accuracy and robustness, effectively handling complex queries.

\section*{Limitations}

Our VCoT-SQL framework relies on supervised fine-tuning (SFT) to train both the SQL generator and SQL validator. While SFT provides a solid foundation, it has a key drawback: it teaches the model to mimic the training examples but does not directly optimize for the true goal of producing executable and accurate SQL queries. Recent research has begun to enhance SFT by incorporating reinforcement learning algorithms such as GRPO, which can improve the model using feedback signals. We plan to address this in future work by incorporating reinforcement learning methods, which can use execution results and designed reward signals to directly optimize the model for better performance.

\section*{Acknowledgements}
This work is supported by the National Key Research and Development Program of China under grant 2025YFC3309504 and the National Natural Science Foundation of China under grants 62436003.

\bibliography{custom}

\newpage
\appendix

\section{Schema Filtering Details and Serialization Format}
\label{appendix:schema}

\begin{figure*}
  \centering
  \includegraphics[width=\linewidth]{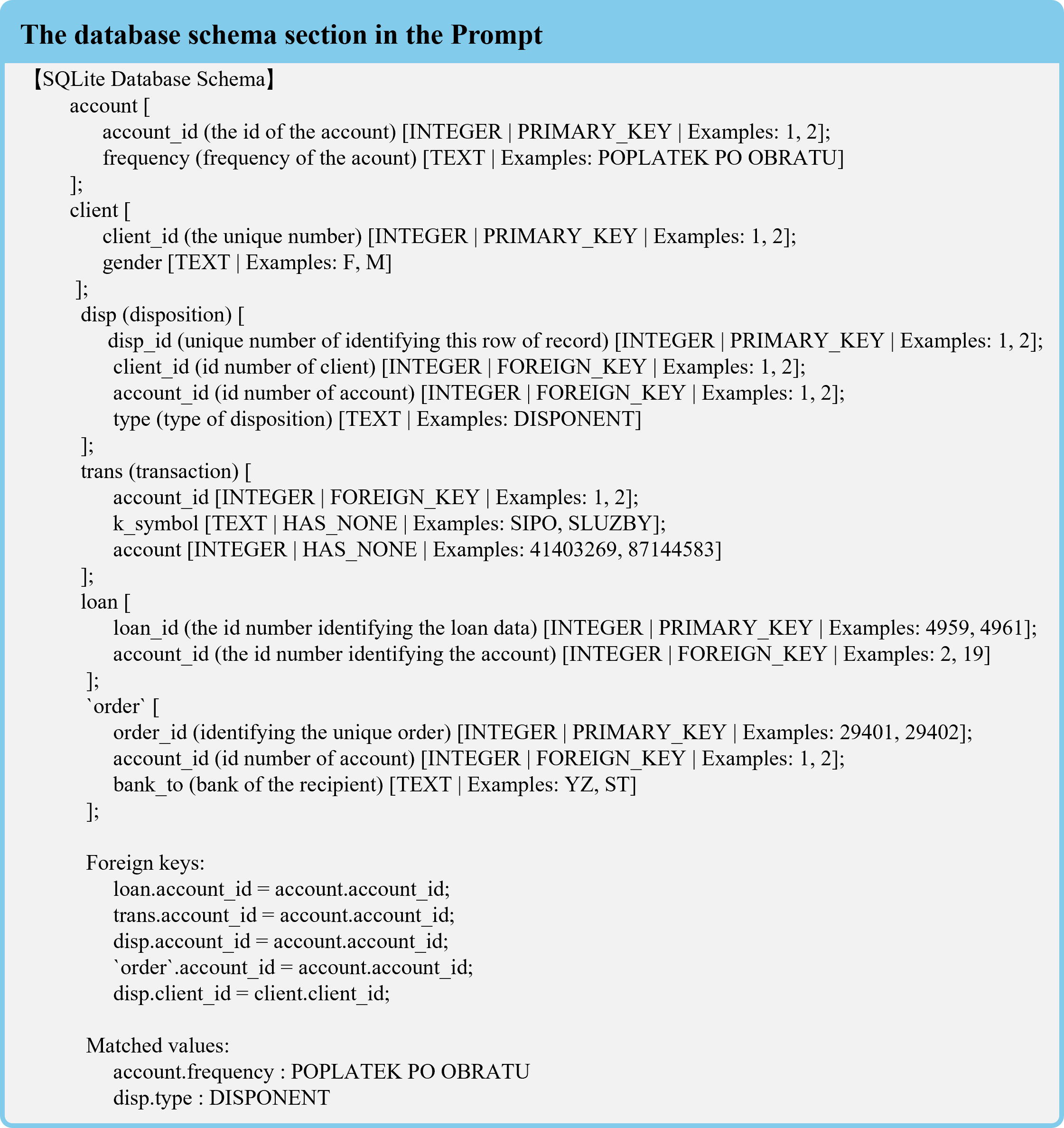}
  \caption{\boldmath{An example of the serialized schema representation used in our framework}}
  \label{schema_serialization}
\end{figure*}

\begin{table*}[t]
\centering
\small
\setlength{\tabcolsep}{5pt}
\begin{tabular}{l|cc|c|ccc}
\toprule
\makecell{Method} & 
\makecell{Spider \\ Dev} & 
\makecell{Spider \\ Test} & 
\makecell{BIRD \\ Dev} & 
\makecell{Spider- \\ DK} & 
\makecell{Spider- \\ Syn} & 
\makecell{Spider- \\ Realistic} \\
\midrule

% --- 第2行 ---
\makecell[l]{only SQL Generator} 
&  85.3	&86.7	&64.86	&76.8	&77.6	&82.7 \\
\midrule

\makecell[l]{Validator FT w/o multi-stage verification} 
& 85.6	&86.7	&65.78	&77.0	&77.9	&83.1\\

% --- 第3行 ---
\makecell[l]{Validator FT w/ multi-stage verification} 
&  85.7	&86.9	&66.49	&77.2	&78.5	&83.3\\
\bottomrule

\end{tabular}
\caption{Effect of multi-stage verification on CoT training data quality.}
\label{tab:CoT_validation}
\end{table*}

\begin{table}[t]
\centering
\small
\renewcommand{\arraystretch}{1.0}

\begin{tabular}{l|l|ccc}
\toprule

\multicolumn{2}{c|}{}
&
\multicolumn{3}{c}{After Validation}
\\

\cmidrule(l){3-5}

\multicolumn{2}{c|}{}
&
Correct
&
Incorrect
&
Total
\\

\midrule

Before
&
Correct
&
64.67
&
0.20
&
64.86
\\

&
Incorrect
&
1.83
&
33.31
&
35.14
\\

Validation
&
Total
&
66.49
&
33.51
&
100
\\

\bottomrule
\end{tabular}

\caption{
Transition matrix of execution correctness (\%) before and after applying the validator on BIRD Dev.
}

\label{tab:validator_robustness}

\end{table}

Regarding the implementation details of the LLM-based components in our schema filter, both the coarse-grained Pre-SQL generation and the keyword extraction processes are performed using the Qwen2.5-Coder-7B-Instruct~\cite{hui2024qwen2} model. This lightweight yet powerful code-centric LLM provides efficient and reliable initial cues for the subsequent retrieval stages.

Furthermore, some datasets provide additional auxiliary information beyond basic table and column names, such as full names of tables and columns, textual descriptions, and external evidence. We leverage this auxiliary information to enable more accurate schema filtering.

When external evidence is available, we combine it with the user's question and extracted keywords as query signals to retrieve relevant columns and values. This allows the retrieval process to capture implicit semantic cues that may not be explicitly mentioned in the question. For the column retriever, when a column is associated with a full name or a textual description, we encode each of these representations separately. The final similarity score between a column and a query (i.e., the question, extracted keywords, or external evidence) is computed as the maximum similarity among the column name, its full name, and its description. This strategy enables more robust matching, especially in cases where abbreviated schema names are insufficient to convey semantic meaning. And in our experiments, we set the top-K of the column retriever to 3, meaning each embedding retrieves at most 3 most relevant columns.

After obtaining the filtered schema elements from the three retrieval paths, we design a structured serialization format to present the schema to the LLM for downstream reasoning and SQL generation.Each table in the serialized schema includes both its table name and full name (if available). The filtered columns are organized in the following order: primary keys, foreign keys, and other columns. For each column, we include the following attributes: the column name, full name, column description, data type, whether it is a primary key, whether it is a foreign key, whether it contains null values, and two representative example values. In addition, all detected foreign key relations and matched values retrieved during the retrieval process are explicitly listed to further guide the model during reasoning and generation. Figure~\ref{schema_serialization} illustrates a concrete example of the serialized schema representation used in our framework.

\section{Experimental Details}
\label{appendix:im_details}
For a more detailed discussion of baseline selection, we compare our method with representative Text-to-SQL systems under comparable inference settings. Some recent works~\cite{sheng2025csc,sheng2025slm} are not included in the main comparison because they adopt a multi-candidate generation and selection strategy with substantially larger inference budgets (from 4 to 64 candidates), whereas our method performs only one generation and one correction step. Under lower-budget settings (4 and 8 candidates), our method achieves better performance, indicating favorable efficiency-performance trade-off under constrained inference budgets.

We describe the construction process of our training datasets for both modules as follows. For the SQL Generator, we use the Spider training set (8,659) and BIRD training set (9,428) to generate CoT data. After applying our CoT quality validation mechanism, we obtain 5,477 and 3,669 high-quality CoT-augmented samples for Spider and BIRD, respectively. For the SQL Validator, we use the validated BIRD training set as a more challenging source, to generate validation CoT data, yielding 947 filtered samples for validator training.

Due to GPU memory limitations, VCoT-SQL was fine-tuned using bfloat16 mixed precision and model sharding, implemented via the integration of the Accelerate framework and DeepSpeed ZeRO-3~\cite{rajbhandari2020zero}. Training was carried out on a server equipped with eight NVIDIA L40 GPUs, and the complete fine-tuning of both models was completed within a few hours.  For each base model, we all set the number of epochs, context length, weight decay, and gradient clipping to 4, 2048, 0.1, and 1.0, respectively. These hyperparameter settings follow OmniSQL~\cite{li2025omnisql} and are commonly adopted in fine-tuning practices.

\section{Validator Data Quality and Robustness Analysis}
\label{appendix:validator}
To better understand the effectiveness of the proposed validator, we further analyze two complementary aspects: the quality of the chain-of-thought (CoT) training data and the reliability of validator behavior during inference. Specifically, we first examine the construction and verification process of CoT supervision, and then evaluate whether the validator introduces reliable corrections while avoiding harmful modifications.

\subsection{Quality Assurance of CoT Examples}
\label{appendix:v_exampler}
To ensure the quality of the validator training data, we employ a hybrid human--LLM data construction pipeline with multi-stage verification mechanisms.

\textbf{1) Error-oriented CoT design.}
The validator’s CoT reasoning process is explicitly structured around six representative SQL error categories that are commonly adopted in prior Text-to-SQL research \cite{shen2025study,xu2025boosting}: syntactic error, column misuse, table misuse, condition errors, aggregation errors, and data type misuse. By constraining the reasoning process to these targeted failure modes, the design improves both the interpretability and consistency of validator decisions while reducing unnecessary reasoning variance.

\textbf{2) Human-curated seed examples.} We construct a small set (fewer than 10) of error-correction CoT demonstrations from real generation errors observed on Spider and BIRD datasets. These examples are manually curated by Text-to-SQL experts to ensure correctness and alignment with realistic error distributions.

\textbf{3) Scalable CoT expansion with verification.} Starting from the curated seed examples, we use LLMs to generate additional CoT training samples. To maintain data quality, all generated samples are filtered using the multi-stage verification pipeline introduced in Section \ref{sec:method_intro2}.

To quantitatively assess the contribution of the verification process, we additionally train validators using CoT data generated both with and without verification filtering. As shown in Table~\ref{tab:CoT_validation}, verification consistently improves execution accuracy, demonstrating that the proposed filtering pipeline effectively enhances the quality of CoT supervision and leads to more reliable validator behavior.

\subsection{Assessment of Validator Reliability}
To assess validator reliability, we analyze how often the validator modifies generator outputs correctly or incorrectly. Using Qwen2.5-Coder-7B-Instruct on BIRD Dev, we compare Generator-only and Generator+Validator settings. As shown in Table~\ref{tab:validator_robustness}, Generator-only outputs are correct on 64.86\% of all examples and incorrect on 35.14\%. After validation, 0.20\% of all examples move from correct to incorrect, whereas 1.83\% move from incorrect to correct. These percentages are measured over the full dataset. Overall, the validator yields a net execution accuracy gain of 1.63\% (1.83\% $-$ 0.20\%), which is consistent with the improvements reported in Table~\ref{tab:SQL_Verification}. These results indicate that the validator rarely introduces detrimental modifications while providing measurable corrective benefits, demonstrating both the robustness and effectiveness of the proposed validation framework.

\begin{table}[t]
\centering
\small
\renewcommand{\arraystretch}{1.0}
\begin{tabular}{l|cc|c}
\toprule
 & Unval. & Val. & Val. Gain \\
\midrule
Syntax-order & 76.13 & 78.36 & +2.23 \\
Logic-order & 76.55 & 78.99 & +2.44 \\
\midrule
Logic-order Gain & +0.42 & +0.63 & +0.21 \\
\bottomrule
\end{tabular}
\caption{2$\times$2 factorial analysis of template ordering and CoT
validation.}
\label{tab:factorial_analysis}
\end{table}
\begin{table*}[t]
\centering
\small
\setlength{\tabcolsep}{5pt}
\renewcommand{\arraystretch}{1.0}
\begin{tabular}{l|cc|c|ccc}
\toprule
Method & \makecell{Spider \\ Dev} & \makecell{Spider \\ Test} & \makecell{BIRD \\ Dev} & \makecell{Spider- \\ DK} & \makecell{Spider- \\ Syn} & \makecell{Spider- \\ Realistic} \\
\midrule
Generator only & 85.3 & 86.7 & 64.86 & 76.8 & 77.6 & 82.7 \\
w/ fine-tuned 7B Validator & 85.7 & 86.9 & 66.49 & 77.2 & 78.5 & \textbf{83.3} \\
w/ GPT-5.4 Validator & \textbf{86.3} & \textbf{87.4} & \textbf{67.99} & \textbf{80.9} & \textbf{79.3} & 82.7 \\
\bottomrule
\end{tabular}
\caption{Upper-bound comparison of validators using outputs from the Qwen2.5-Coder-7B-Instruct Generator.}
\label{tab:frontier_validator}
\end{table*}

\begin{table*}[t]
\centering
\small
\setlength{\tabcolsep}{6pt}
\renewcommand{\arraystretch}{1.0}
\begin{tabular}{l|ccc|c}
\toprule
Backbone & DB perturbation & NLQ perturbation & SQL perturbation & Average \\
\midrule
DeepSeek-Coder-6.7B-Instruct & 86.2--69.6 & 85.8--74.2 & 84.3--82.3 & 85.4--75.8 \\
Qwen2.5-Coder-7B-Instruct & 85.0--72.5 & 84.8--74.9 & 85.7--81.7 & 85.1--76.5 \\
XiYanSQL-QwenCoder-7B & \textbf{87.0--73.5} & \textbf{87.3--76.6} & \textbf{86.6--84.2} & \textbf{87.1--78.3} \\
\bottomrule
\end{tabular}
\caption{VCoT-SQL robustness on Dr.Spider (pre-/post-perturbation EX).}
\label{tab:drspider}
\end{table*}

\textcolor{black}{\section{Additional Ablation Analysis}}
This section presents additional analyses of validation effectiveness and method robustness.
\subsection{Factorized Analysis of CoT Validation}
\label{app:factorial}
To disentangle the effects of template ordering and CoT validation, 
we perform a 2$\times$2 factorial analysis over the four corresponding 
settings in Table~\ref{tab:CoT_Synthesis}. Table~\ref{tab:factorial_analysis} shows that CoT validation consistently improves EX under both template orders, with gains of 2.23\% for syntax-order and 2.44\% for logic-order. Logic-order improves over syntax-order by 0.42\% without validation and 0.63\% with validation. The resulting interaction effect is only 0.21\%, suggesting that the benefit of CoT validation is largely independent of template ordering.

\subsection{Frontier Validator}
\label{app:frontier_validator}
To estimate SQL-validation headroom, we replace the Validator with GPT-5.4 while fixing the Qwen2.5-Coder-7B-Instruct Generator outputs. This isolates validation. As shown in Table~\ref{tab:frontier_validator}, GPT-5.4 improves over the generation-only baseline on five of the six benchmarks and achieves the largest gain on Spider-DK. It also outperforms our fine-tuned 7B Validator on most datasets. The macro-averaged EX is 79.68\% for the 7B Validator and 80.77\% for GPT-5.4, with an absolute gap of 1.08\%.

\subsection{Dr. Spider Robustness}
\label{app:drspider}
Following the official Dr.Spider~\cite{chang2023drspider} protocol, we evaluate VCoT-SQL on all 17 perturbation test sets: three database (DB), nine natural-language-question (NLQ), and five SQL perturbations. Each test set contains aligned pre- and post-perturbation examples. In Table~\ref{tab:drspider}, $x$--$y$ denotes macro-averaged EX on the corresponding pre- and post-perturbation examples. Across all three backbones, SQL perturbations cause the smallest drops, while DB perturbations cause the largest drops, with NLQ perturbations in between. XiYanSQL-QwenCoder achieves the best overall performance (87.1/78.3). DB perturbations directly alter schema representations and can therefore introduce schema-linking errors that propagate to SQL generation, whereas SQL perturbations generally preserve the question--schema correspondence.

\section{CoT Data Synthesis Prompt Templates}
In this section, we present the prompts used for generating our CoT dataset.

\subsection{Prompt for SQL Generation}
\begin{figure*}
  \centering
  \includegraphics[width=\linewidth]{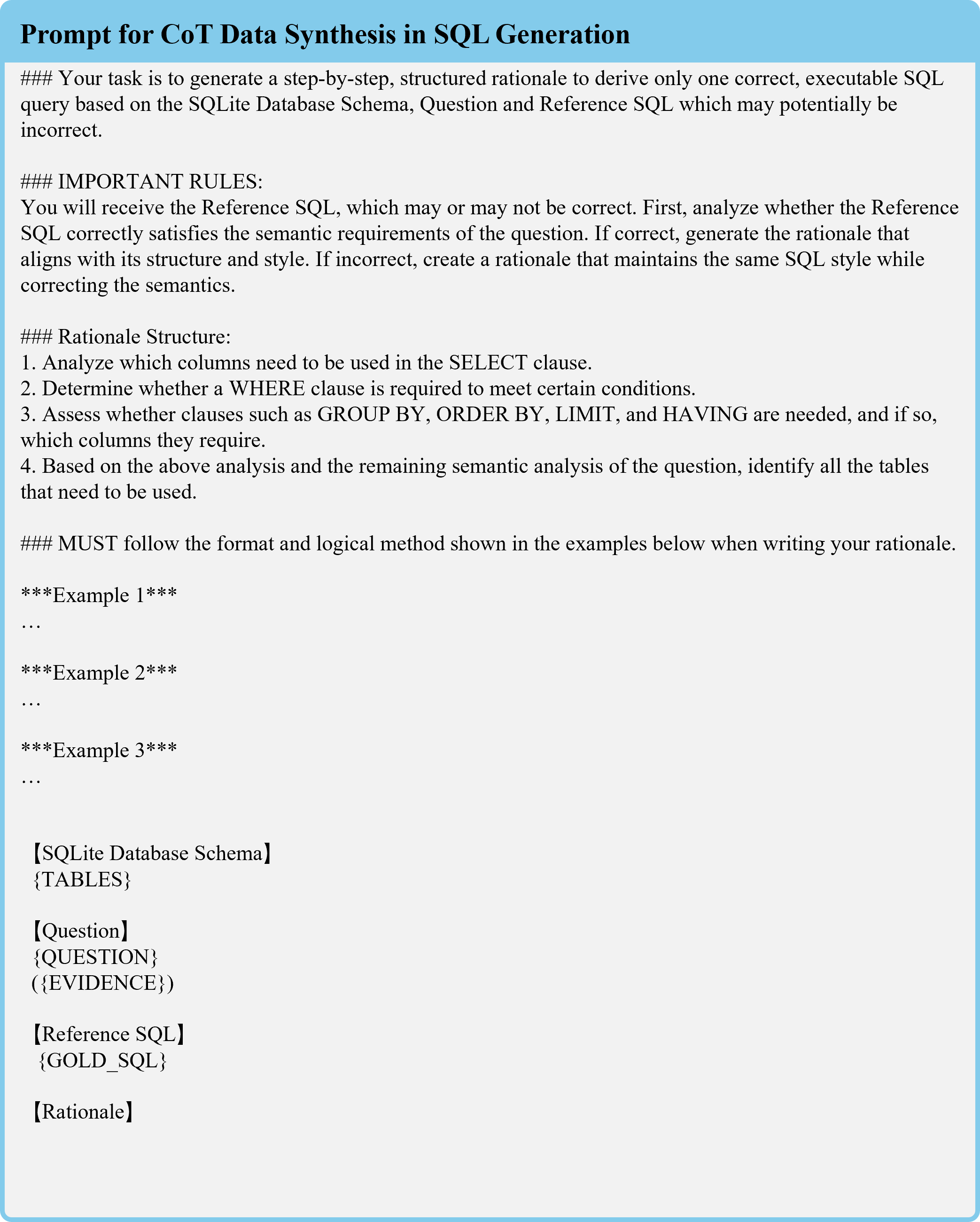}
\caption{The prompt template for synthesizing clause-level CoT rationales in SQL generation.}
\label{prompt_for_sql_generation}
\end{figure*}

\begin{figure*}
  \centering
  \includegraphics[width=\linewidth]{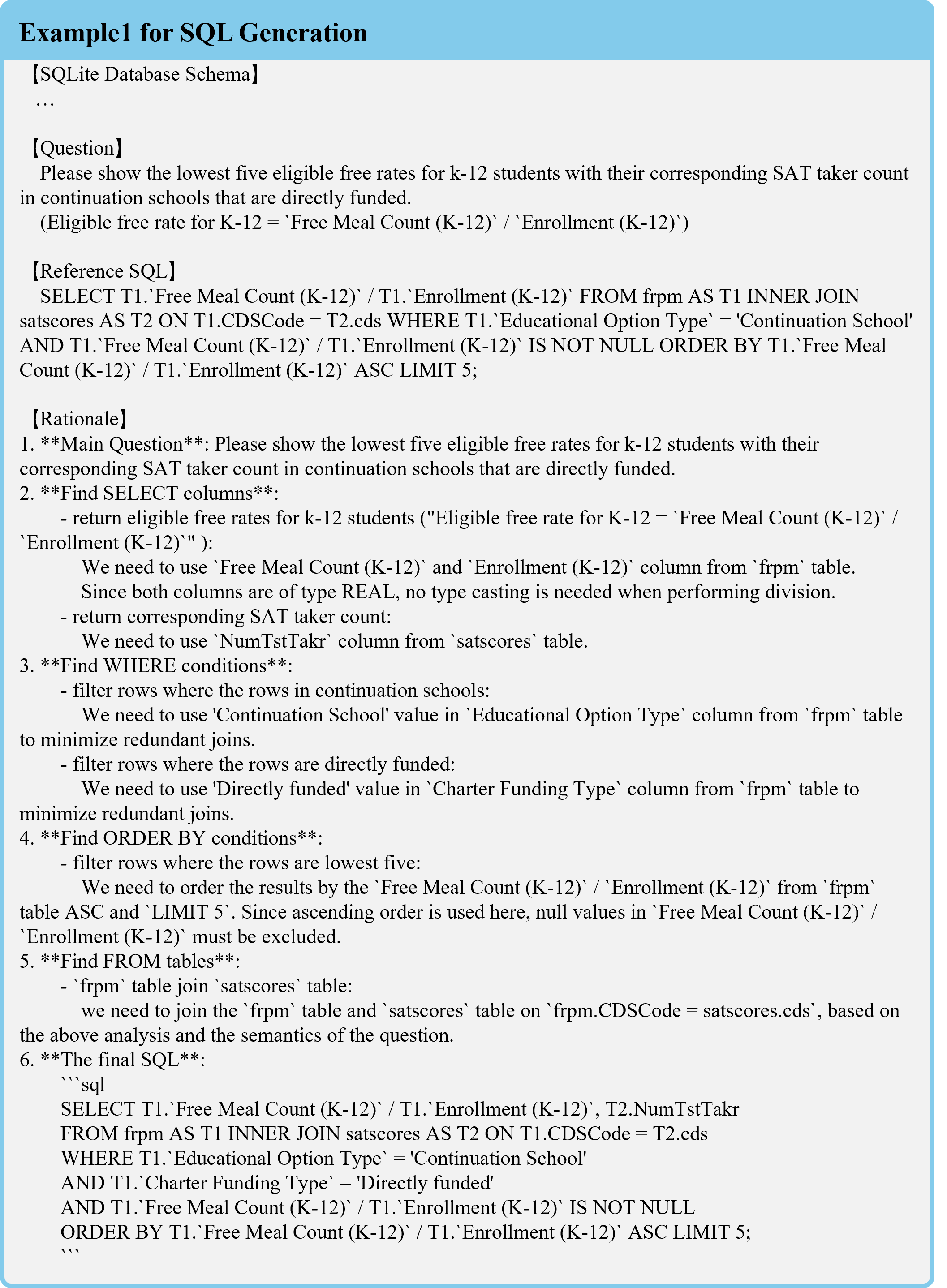}
\caption{The example 1 for synthesizing clause-level CoT rationales in SQL generation.}
\label{example1_for_sql_generation}
\end{figure*}

\begin{figure*}
  \centering
  \includegraphics[width=\linewidth]{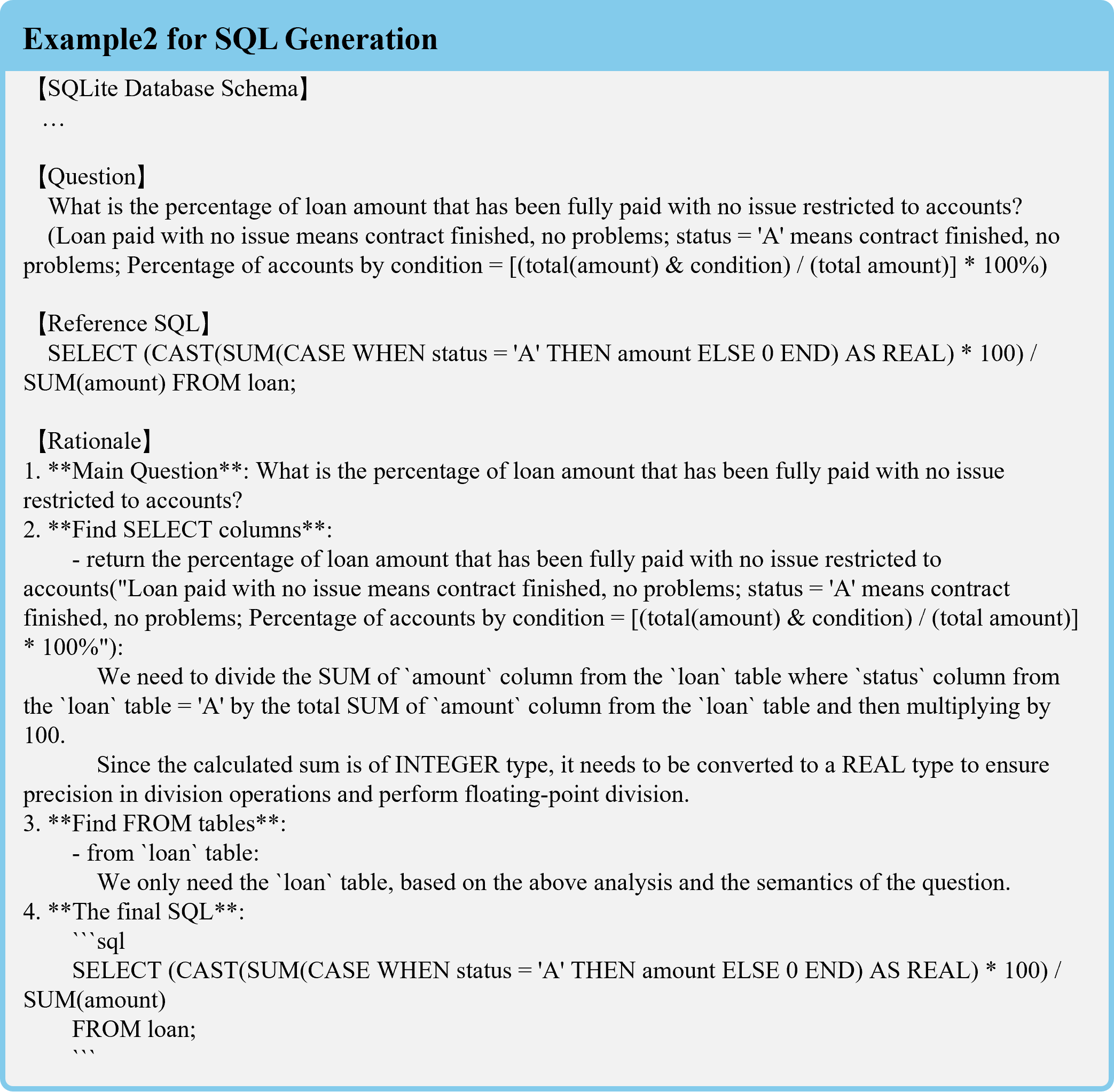}
\caption{The example 2 for synthesizing clause-level CoT rationales in SQL generation.}
\label{example2_for_sql_generation}
\end{figure*}

\begin{figure*}
  \centering
  \includegraphics[width=\linewidth]{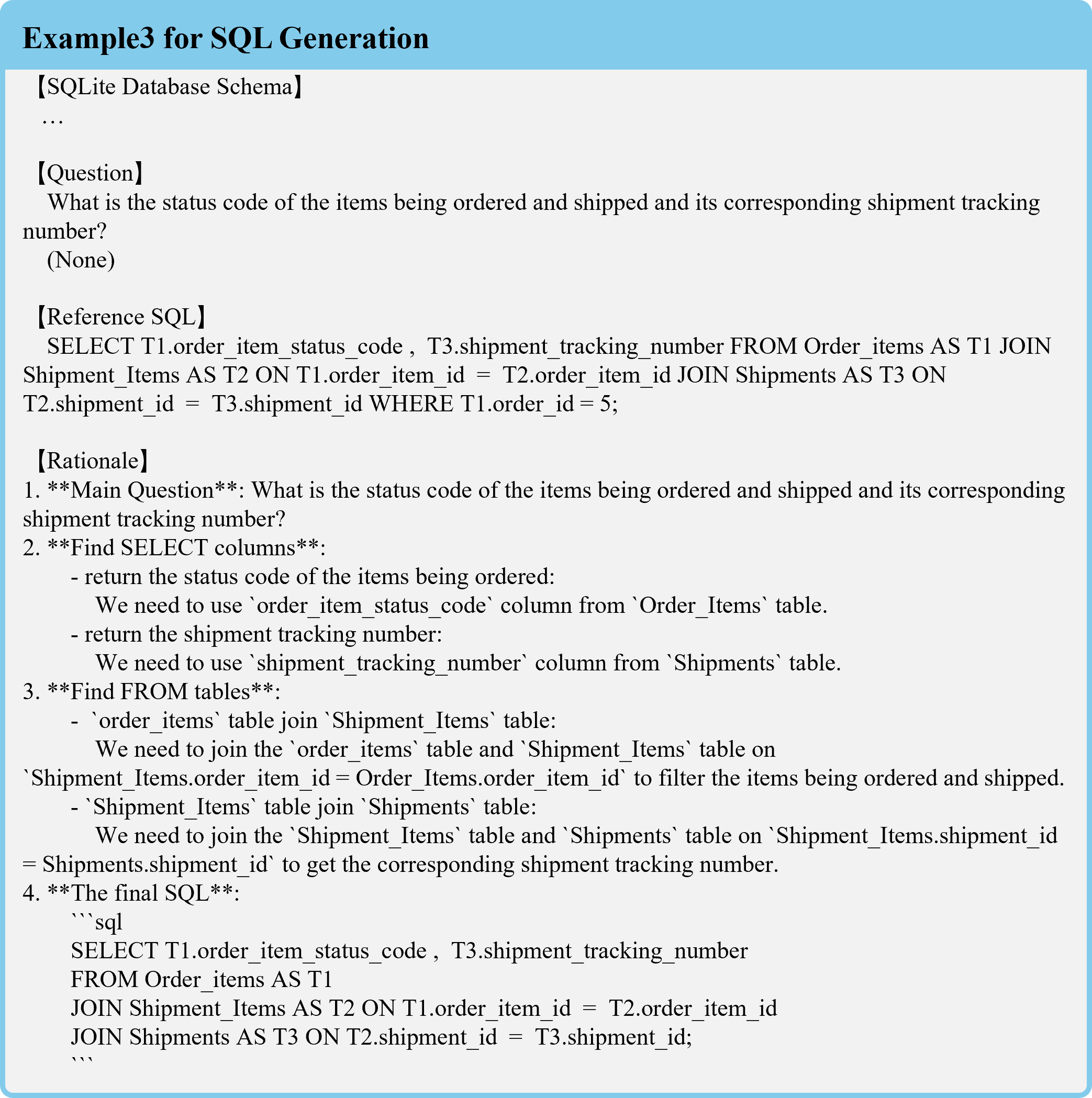}
\caption{The example 3 for synthesizing clause-level CoT rationales in SQL generation.}
\label{example3_for_sql_generation}
\end{figure*}

In our prompt, we use instructions to decompose the generation process into clause-level steps: SELECT, WHERE, GROUP BY/ORDER BY/LIMIT/HAVING, and FROM, and provide 3 examples as few-shot demonstrations to guide Qwen2.5-Coder-7B-Instruct in generating intermediate reasoning steps. The prompt used for synthesizing the generation CoT is illustrated in Fig ~\ref{prompt_for_sql_generation}, and the three few-shot examples are presented in Fig ~\ref{example1_for_sql_generation}, Fig ~\ref{example2_for_sql_generation} and Fig ~\ref{example3_for_sql_generation} .
\label{appendix:sql_generation_prompt}

\subsection{Prompt for SQL Correction}
\label{appendix:sql_validation_prompt}
\begin{figure*}
  \centering
  \includegraphics[width=\linewidth]{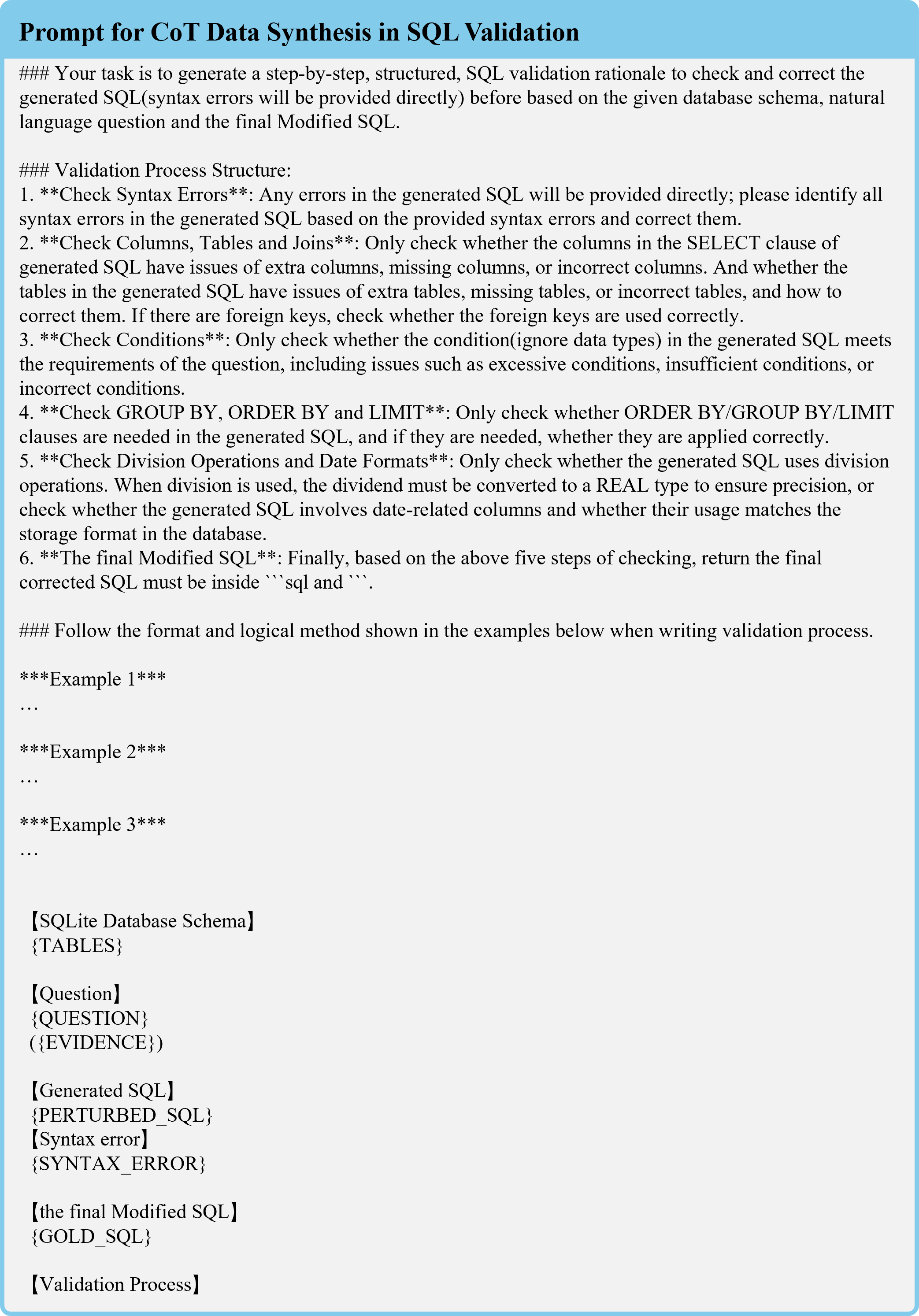}
\caption{The prompt template for synthesizing multi-type CoT rationales in SQL validation.}
\label{prompt_for_sql_validation}
\end{figure*}

\begin{figure*}
  \centering
  \includegraphics[width=\linewidth]{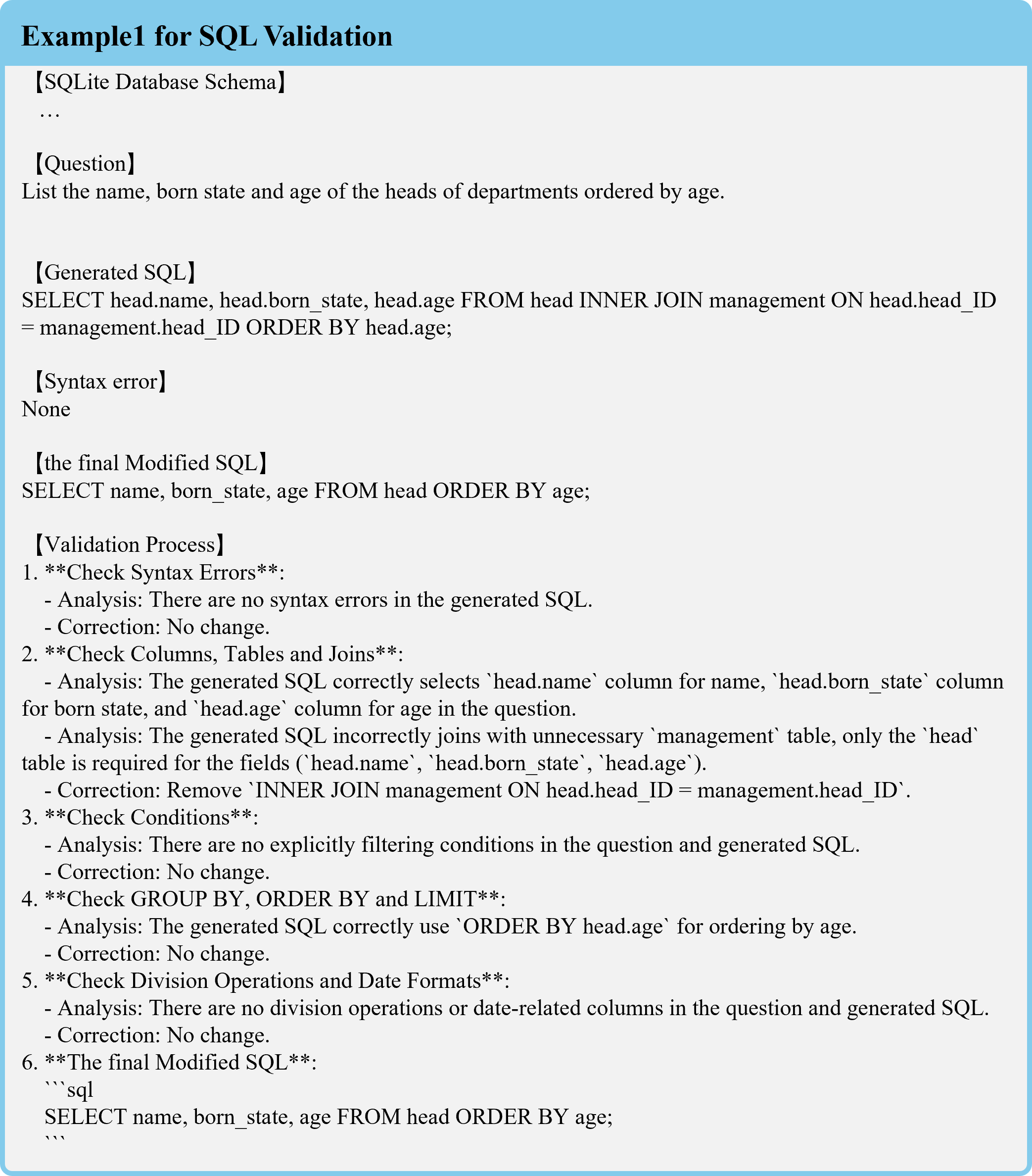}
  \caption{The example 1 for synthesizing multi-type CoT rationales in SQL validation.}
\label{example1_for_sql_validation}
\end{figure*}

\begin{figure*}
  \centering
  \includegraphics[width=\linewidth]{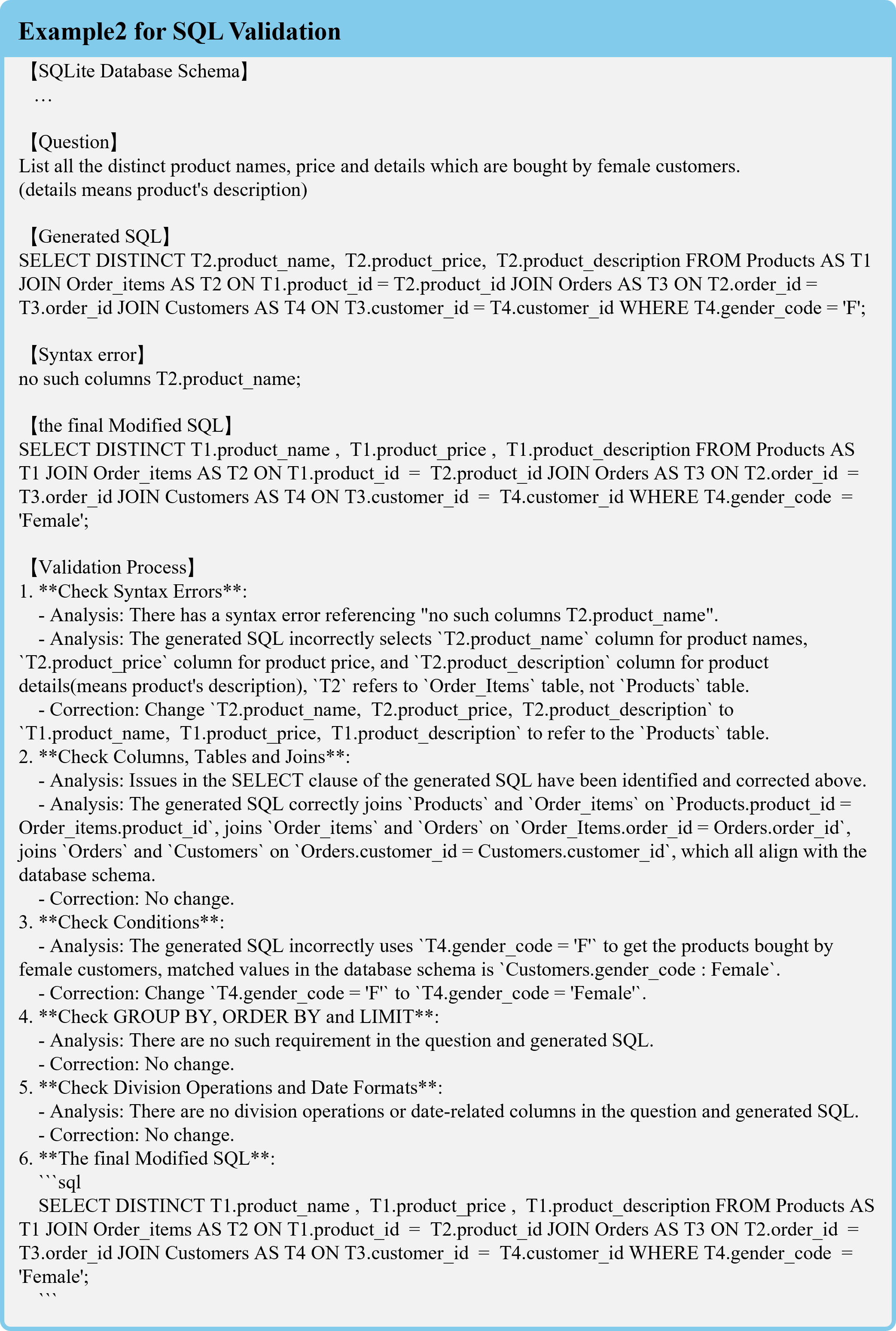}
    \caption{The example 2 for synthesizing multi-type CoT rationales in SQL validation.}
\label{example2_for_sql_validation}
\end{figure*}

\begin{figure*}
  \centering
  \includegraphics[width=\linewidth]{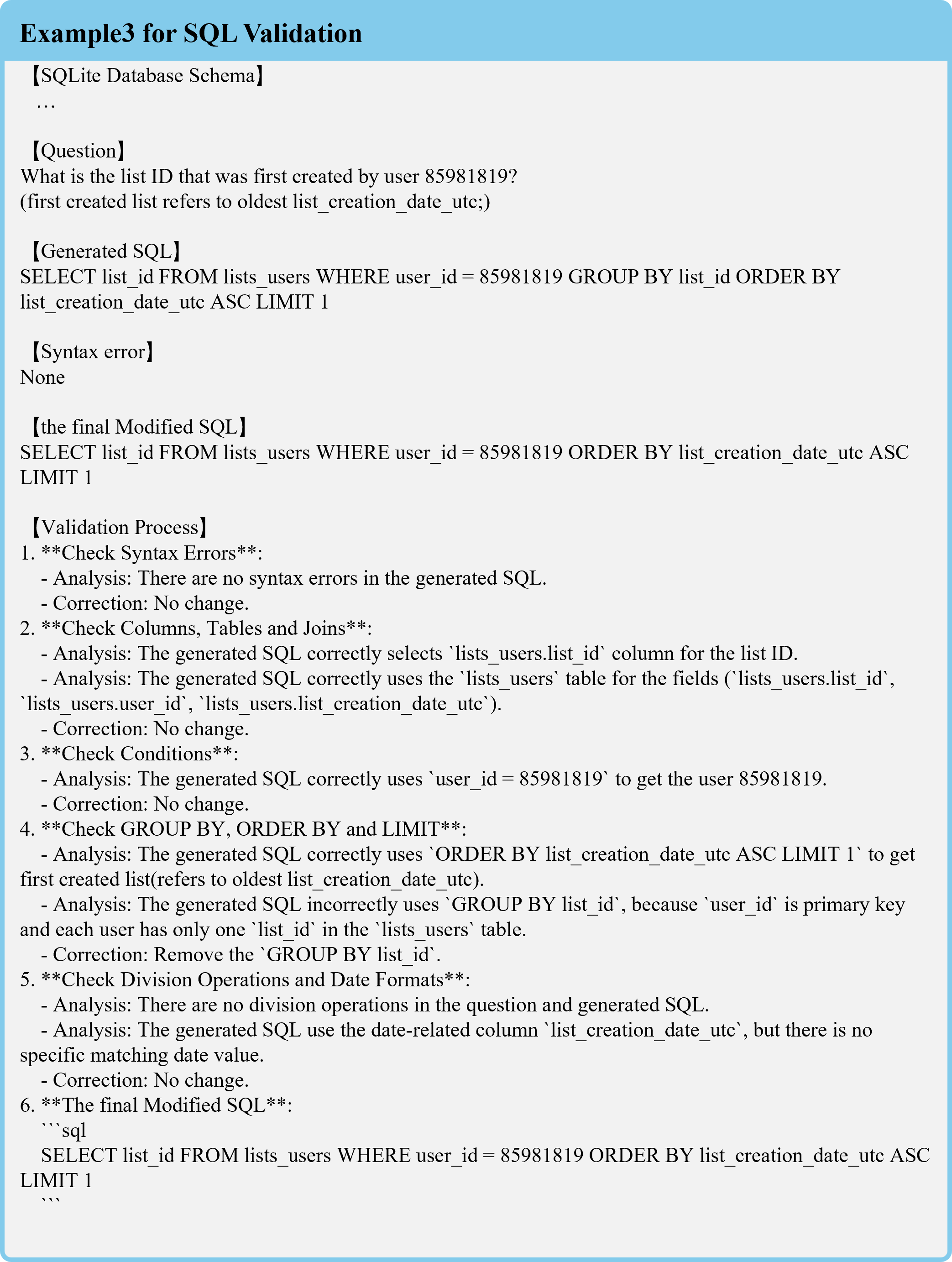}
    \caption{The example 3 for synthesizing multi-type CoT rationales in SQL validation.}
\label{example3_for_sql_validation}
\end{figure*}

In our prompt, we use instructions to decompose the validation process into multi-type checking steps: Check Syntax Errors, Check Columns, Tables and Joins, Check Conditions, Check GROUP BY/ORDER BY/LIMIT, Check Division Operations and Date Formats, and provide 3 examples as few-shot demonstrations to guide Qwen2.5-Coder-7B-Instruct in generating intermediate validation reasoning steps. The prompt used for synthesizing the validation CoT is illustrated in Fig.~\ref{prompt_for_sql_validation}, and the three few-shot examples are presented in Fig.~\ref{example1_for_sql_validation}, Fig.~\ref{example2_for_sql_validation} and Fig.~\ref{example3_for_sql_validation}.

%--- 附录 B：推理 ---
\section{Inference Prompt Templates}
\label{sec:inference_prompts}
\begin{figure*}
  \centering
  \includegraphics[width=\linewidth]{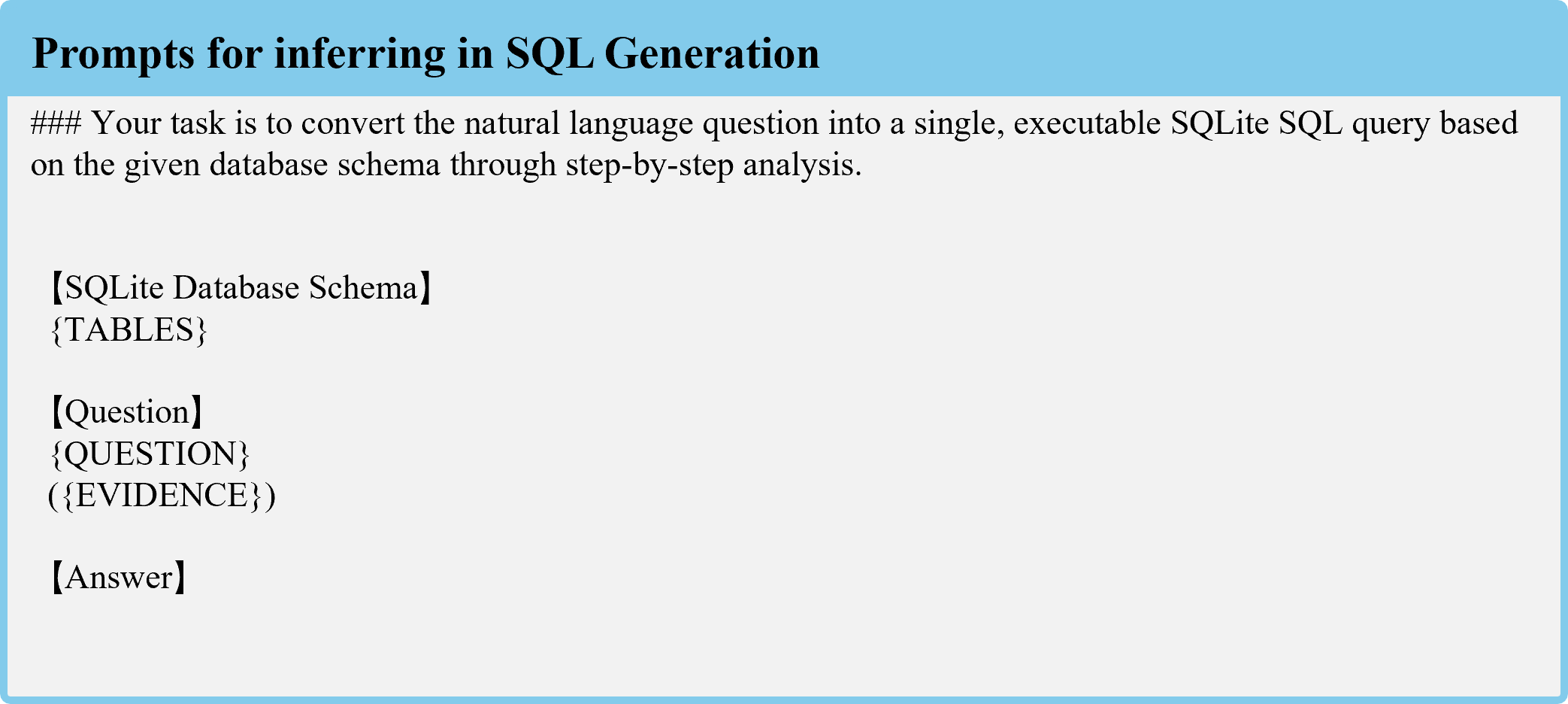}
\caption{Inference prompt template for the SQL generation stage.}
\label{prompt_for_sql_generation_inference}
\end{figure*}

\begin{figure*}
  \centering
  \includegraphics[width=\linewidth]{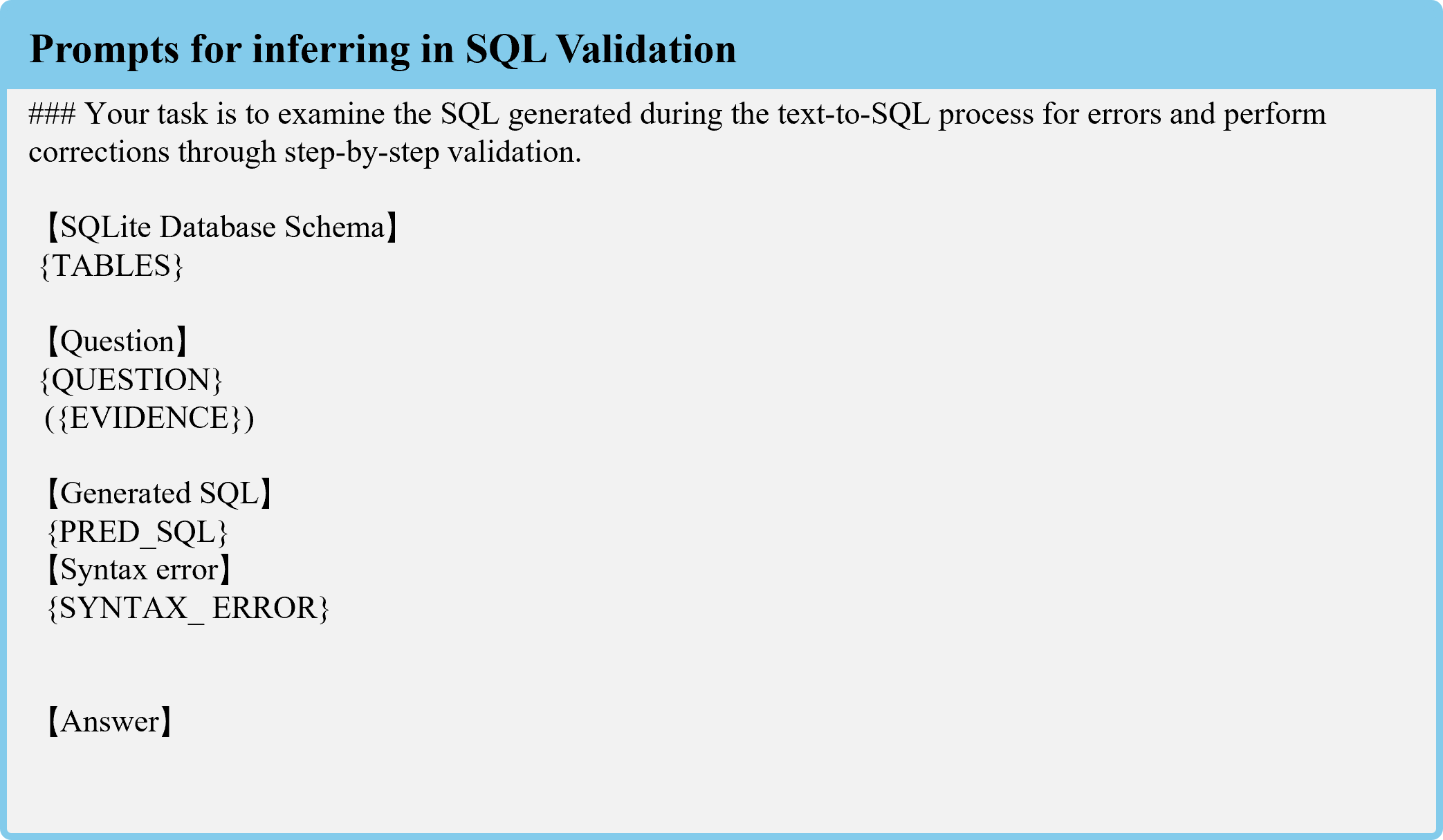}
\caption{Inference prompt template for the SQL validation stage.}
\label{prompt_for_sql_validation_inference}
\end{figure*}

For the inference stage, we employ the concise prompts for SQL generation and validation, as shown in Fig ~\ref{prompt_for_sql_generation_inference} and Fig ~\ref{prompt_for_sql_validation_inference}, respectively.

\end{document}